\documentclass[12pt]{article}

\usepackage{sbc-template}
\usepackage{graphicx,url}
\usepackage[utf8]{inputenc}

\usepackage{glossaries}
\usepackage{booktabs}
\usepackage{multirow}
\usepackage[normalem]{ulem}
\usepackage{xcolor}
\usepackage{ifthen}

\newif\ifarxiv
\arxivtrue

\usepackage{transparent}
\usepackage{tikz}
\ifarxiv
    \newcommand\copyrighttext{%
      \scriptsize Accepted for publication at SIBGRAPI WIP 2026. The final version will be published at SBC-OpenLib (SOL).}
    \newcommand\copyrightnotice{%
    \begin{tikzpicture}[remember picture,overlay]
    \node[anchor=south,yshift=30pt,xshift=0pt] at (current page.south) {\fbox{\transparent{0.85}\mbox{\copyrighttext}}};
    \end{tikzpicture}%
    }
\else
\fi

\title{VeriCam: A Verification Baseline for the Classification of Unknown Data}

\author{Lucas Wojcik\inst{1}, Gabriel E. Lima\inst{1}, Sergio M. Silva Jr.\inst{1}, \\Eduil Nascimento Jr.\inst{2}, David Menotti\inst{1}}

\address{Department of Informatics, Federal University of Paran\'a, \\ Curitiba, Brazil
\nextinstitute
Department of Technological Development and Quality, \\Paran\'{a} Military Police, Curitiba, Brazil
}

\begin{document}

\newacronym{alpr}{ALPR}{Automatic License Plate Recognition}
\newacronym{ood}{OOD}{Out of Distribution}
\newacronym{dl}{DL}{Deep Learning}
\newacronym{vit}{ViT}{Vision Transformer}
\newacronym{nlp}{NLP}{Natural Language Processing}

\maketitle

\ifarxiv
    \copyrightnotice
\else
\fi

\ifarxiv
  \vspace{-3mm}
\else
\fi

\begin{abstract}
The advent of foundation models has enabled a new era in zero-shot classification.
Yet, key challenges persist.
Despite their impressive generalization power that leverages the immense pre-training knowledge, both foundation models for image and text, as well as vision-text hybrids, lack the representational power needed for fine-grained, minutiae-based class separation that some real-world tasks require.
To address the current gaps in the literature, we propose VeriCam, a pipeline designed to learn highly specialized features that enable the classification of unknown classes in unseen data.
VeriCam works by leveraging the representation power of image models trained for the verification task, where the model develops an intricate feature space that incorporates fine-grained details.
By training a model to discriminate between pairs of images from the same or different classes, a relational graph is constructed, representing the class relationships between data points.
We then present two approaches for graph clustering: a naive algorithm and a specific setup for the Leiden graph clustering algorithm.
The pipeline is validated on the LPLCv2 dataset, which comprises real-world traffic surveillance images.
We show that the dataset carries an inherent capture device bias that poses a generalization challenge for downstream License Plate recognition tasks such as OCR.
As such, we dynamically identify capture devices with a label-agnostic approach, enabling the construction of a fair and unbiased benchmark.
In the cross-device scenario, our pipeline reaches an F1-Score of 93.45 in the verification baseline and a V-Measure score of 80.13 in the clustering step.
All code is publicly available at \url{https://github.com/lmlwojcik/VeriCam}

\end{abstract}

\section{Introduction}

\gls{dl} has established itself as the state-of-the-art approach for automatic pattern recognition \textit{par excellence}, a development enabled by the large amounts of data available today~\cite{zha2025data}.
Owing to its data-driven nature, its success often depends on the quantity and quality of training data~\cite{hoffmann2022training}.
Each real-world application of \gls{dl}, therefore, sees its effectiveness inherently linked to the quality of the training data available, often leveraging domain-specific knowledge encoded in the annotated labels.

As such, one of the major real-world challenges for the state of the art consists of dealing with inconsistent, incomplete or misleading data.
Some domain-specific constraints may also render some tasks unmanageable for classic approaches.
In particular, we are interested in the case where the number of classes is unknown and unconstrained a priori.
This scenario cannot be handled through standard classification models, as these employ a fixed number of output neurons to encode an already known number of classes.

A few strategies have been employed over the years to deal with this issue.
Early research developed \gls{ood} classification~\cite{hendrycks2017baseline}, which extends the standard \(N\) classification to an \(N+1\) scenario, where the model should also be capable of identifying instances not belonging to the initial \(N\) classes.
Some recent works extend \gls{ood} towards zero-shot classification~\cite{yin2019benchmarking}, aiming to categorize unseen data into new classes, effectively tackling the multi-unknown class issue.
However, these approaches face significant limitations represented by overall lower accuracy and the complexity of domain-specific knowledge, limiting their~generalization potential~\cite{vaze2022gcd,wen2023parametric}.

An important application of zero-shot classification, and the one to which we apply our proposal, is capture device recognition in \gls{alpr} datasets.
A previous \gls{alpr} study has unraveled unexpected biases in street surveillance datasets~\cite{laroca2022first}, showing that a very small CNN can correctly identify the datasets that each traffic image instance comes from when trained on a standard classification task.
While this largely impacts cross-dataset generalization, intra-dataset contamination can also be seen in some large datasets.
We show that intra-dataset experiments can also be skewed by an invisible bias in the form of device contamination in the testing sets.
As such, device recognition is an important step towards building a fair, unbiased benchmark.

With these motivations in mind, we propose VeriCam, a \gls{dl} pipeline designed to enable the classification of unknown classes.
Instead of mapping instances to a predefined label space, we reframe zero-shot inference using a pairwise binary decision task.
We employ a verification network to estimate the probability that two samples belong to the same latent class, and build a graph representation of the target dataset from which same-class clusters can be identified and extracted.
By decoupling class discrimination from static semantic signatures, we learn a universal, domain-invariant feature space to determine semantic identity spanning novel classes.
In this work, we narrow our focus on validation within the street surveillance domain, tailoring the approach for a practical application.

Therefore, our contributions can be summarized as:

\begin{itemize}
    \item A novel method for zero-shot classification based on a verification network.
    \item Two algorithms that employ the original intuition, and their experimental validation.
    \item A short study on capture device contamination for OCR efficiency.
    \item The open source implementation of all pipeline steps, available at \url{https://github.com/lmlwojcik/VeriCam}.
\end{itemize}

The remainder of this paper is organized as follows.
Section~\ref{sec:related} presents related work and the current state of the art behind our motivation and proposal.
Section~\ref{sec:method} details our proposed pipeline and algorithms, as well as the target dataset.
Section~\ref{sec:experiments} details the experiments used to validate our approach.
Section~\ref{sec:results} presents the results for each experimental scenario.
Finally, Section~\ref{sec:conclusion} concludes the paper.

\section{Related Work}
\label{sec:related}

Recent literature has tackled unknown data as an \gls{ood} problem~\cite{yang2024survey}, which consists of detecting data points that are not represented in known classes.
One of the main approaches in this task is to threshold the output logit map of the network, rejecting low confidence detections~\cite{liu2020energy,zhang2023decoupling}.
Other methods include distribution analysis and prototype synthesis~\cite{lu2025out}.

The main drawback of simple \gls{ood} is the inability to discover and classify unknown classes.
In this sense, zero-shot classification is a generalization of \gls{ood}, such that the model must be able to recognize and classify unseen classes at test time.
Zero-shot learning is often studied in the \gls{nlp} field~\cite{yin2019benchmarking,saha2024text}, where LLMs are widely used~\cite{zhang2024generation}.

The text-based zero-shot classification methods, however, are not directly transferable to the image domain.
While the large training corpus of an LLM provides enough vocabulary to enable generalization of novel sentence arrangements, these models face a notorious hurdle when dealing with domain-specific data~\cite{ling2025domain}.
This limitation can also be seen in the visual domain, where applications for zero-shot learning often rely on highly specialized features, such as the device detection task tackled in this work.

Despite this limitation, image foundation models are still used for zero-shot image classification.
Approaches vary from training with synthetically generated data~\cite{shipard2023synthetic} to leveraging text-based descriptions~\cite{saha2024text,novack2023chils}.
In both cases, methods are limited by the general knowledge of the models employed for knowledge extraction.

Other methods of zero-shot classification may leverage knowledge acquired at training time, such as features shared across known classes~\cite{mensink2014costa}.
The COSTA framework, for example, explicitly models knowledge transfer by extracting and leveraging the statistical co-occurrences between classes.
Other models use class prototypes, utilizing the learned latent embedding space as a source of feature detection in order to define new classes from bits and pieces of the detected features~\cite{xian2016latent}.

These approaches, however, are also limited in terms of their applicability.
The main drawback of existing methods in the literature is the lack of a clear way to bound and separate classes that may be similar, but subtly different, such that highly specialized knowledge may be needed in order to tell instances apart.

We address these gaps in the literature by devising a robust method for the classification of unknown classes without the need for a priori knowledge.
The proposed pipeline aims to utilize local comparisons, avoiding globally defined knowledge in favor of proximity features that encode class-based information in their relational similarity metric as opposed to crisp label~values.

\section{Methodology}
\label{sec:method}

Our method exploits local relationships between instances as opposed to global cues and features.
Intuitively, the set of known classes can be built by comparing instances to one another, grouping instances with high similarity and separating instances with low similarity.
In doing so, we draw influence from the facial recognition state of the art~\cite{du2022elements}, which has established the verification task as a foundational method to train feature descriptors~\cite{khalid2023facial}.
It leverages the task's potential to teach the model to separate similar-but-different classes, enabling the learning of robust feature descriptors.
While in facial recognition these models are later used for identification (classification) tasks by swapping the last layer, here we simply use the final model as a feature descriptor for graph~generation.

\subsection{Proposed Pipeline}
\label{sec:proposed}

The proposed pipeline is based on a label-agnostic zero-shot clustering approach driven by the recognition of pairwise relations.
It consists of a feature descriptor model (here, we use the \gls{vit}~\cite{dosovitskiy2021vit}) trained on the verification task (given two images, determine whether they belong to the same class or not) and a graph clustering step.

The model is first trained to learn effective feature descriptors for its specific domain, optimizing for the cosine distance between images of the same class.
For this, we employ the Triplet loss, using the cosine distance defined in Equation~\ref{eq:cos_dist} for feature vectors $V_1$ and $V_2$.
The cosine distance ranges from $2$ for completely different vectors to $0$ for equal vectors, and is based on the cosine similarity, defined in Equation~\ref{eq:cos_sim}.
The similarity, accordingly, ranges respectively from $-1$ to $1$.

\begin{equation}
    CosDist(A,B) = 1 - CosSim(V_1,V_2)
\label{eq:cos_dist}
\end{equation}

\begin{equation}
    CosSim(A,B) = \frac{\mathbf{V_1} \cdot \mathbf{V_2}}{\|\mathbf{V_1}\| \|\mathbf{V_2}\|}
\label{eq:cos_sim}
\end{equation}

\begin{figure*}[!htbp]
\centering
    \begin{minipage}{0.245\linewidth}
        \centering
        {\footnotesize (a) Device 30} \\[0.75mm] %
        \includegraphics[width=0.95\linewidth]{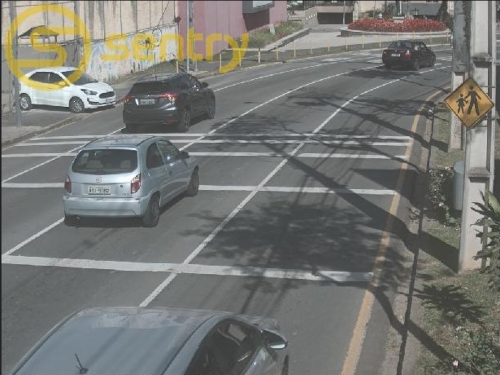} \\[1mm] %
        \includegraphics[width=0.95\linewidth]{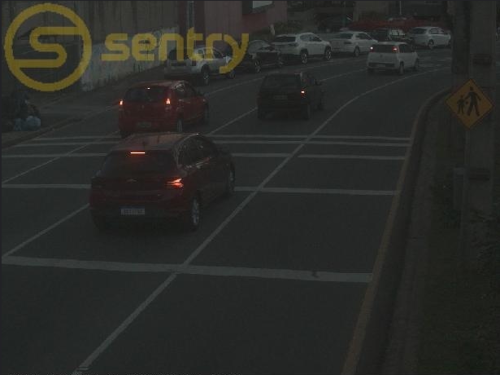} \\[1mm]
        \includegraphics[width=0.95\linewidth]{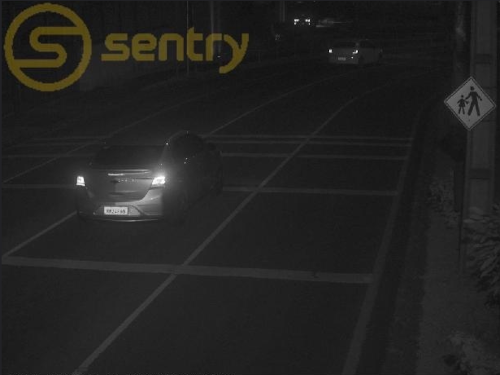} \\[1mm]
    \end{minipage}
    \begin{minipage}{0.245\linewidth}
        \centering
        {\footnotesize (b) Device 48} \\[0.75mm] %
        \includegraphics[width=0.95\linewidth]{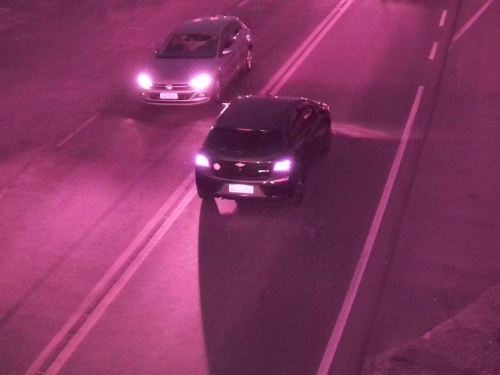} \\[1mm] %
        \includegraphics[width=0.95\linewidth]{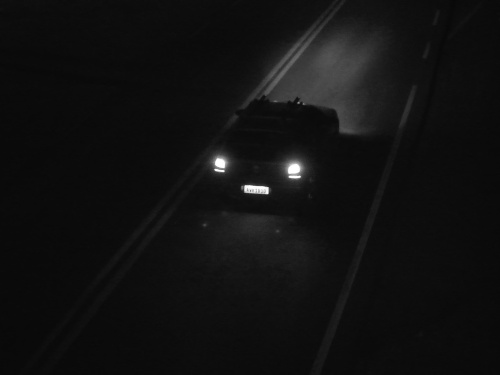} \\[1mm]
        \includegraphics[width=0.95\linewidth]{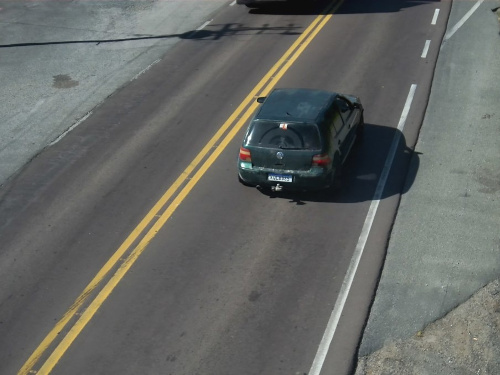} \\[1mm]
    \end{minipage}
    \begin{minipage}{0.245\linewidth}
        \centering
        {\footnotesize (c) Device 468} \\[0.75mm] %
        \includegraphics[width=0.95\linewidth]{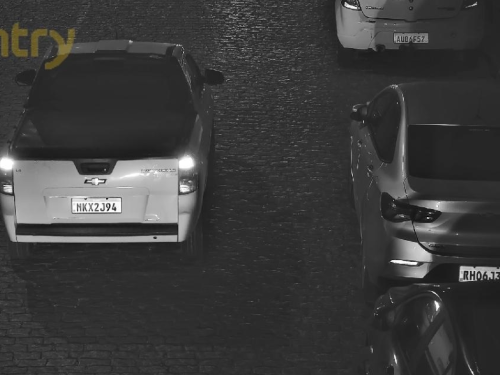} \\[1mm] %
        \includegraphics[width=0.95\linewidth]{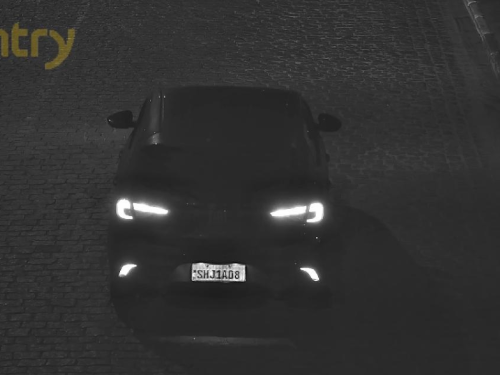} \\[1mm]
        \includegraphics[width=0.95\linewidth]{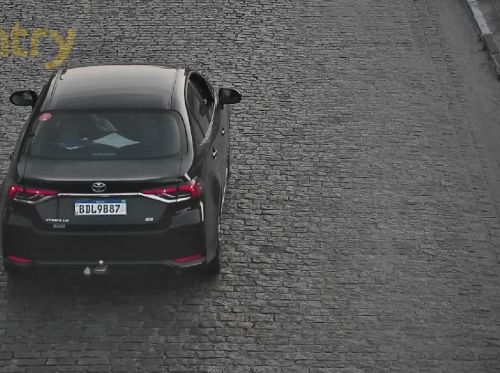} \\[1mm]
    \end{minipage}
    \begin{minipage}{0.245\linewidth}
        \centering
        {\footnotesize (d) Device 546} \\[0.75mm] %
        \includegraphics[width=0.95\linewidth]{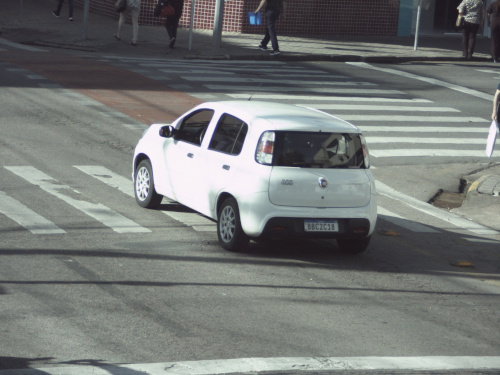} \\[1mm] %
        \includegraphics[width=0.95\linewidth]{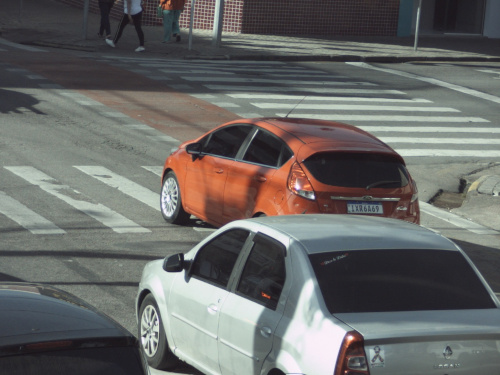} \\[1mm]
        \includegraphics[width=0.95\linewidth]{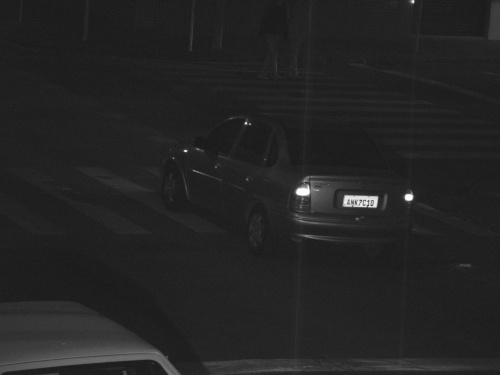} \\[1mm]
    \end{minipage}
    \caption{Instances from the LPLCv2 Dataset.}

\label{fig:instances}
\end{figure*}

Then, the model is used as a feature extractor in order to build a graph-based representation of the instance space that represents their pairwise similarity.
We synthesize local pairwise decisions into a global representation through the affinity matrix $\mathbf{A}$, which represents an undirected graph $G = (V, E)$, where vertices $V$ represent instances and edge weights $E$ denote the cosine similarity between the two vertices it connects.
A graph clustering algorithm is then executed on $G$, yielding new class labels for each test instance.

For the clustering step, our first approach relies on a naive algorithm.
The algorithm starts its execution with a known set of instances and classes, which may be either initialized from data known a priori or left blank and then initialized by the first instance seen.
Then, for each new instance, its cosine similarity to all known images is computed and the instance is incorporated into the known set.
This new instance either initializes a novel class if the mean similarity to all images from each class is less than a predetermined threshold, or assigned to the class with highest mean similarity otherwise.

Our second approach relies on the Leiden algorithm~\cite{traag2019leiden} for graph clustering.
Since our target dataset features a large number of images (over six thousand just for the testing set), spectral clustering~\cite{vonluxburg2007spectral} appears to be prohibitively expensive, and therefore we rely on the Leiden algorithm for its highly efficient heuristics.
Given the nature of our problem, we use the Constant Potts Model (CPM)~\cite{felipe2025leiden} for community detection.
In both cases, this step is treated as a clustering task, given its label-agnostic nature.

\subsection{The Dataset}
\label{sec:dataset}

We utilize the LPLCv2 dataset~\cite{wojcik2026lplcv2}, which is comprised of $37,099$ images, of which $34,760$ are annotated with regards to the camera ID\footnote{This value follows the latest release of LPLCv2.}.
Each camera is defined according to its installation location. As such, each individual device is identified by the scene displayed in the images.
Some examples of this can be found in Figure~\ref{fig:instances}.
Our goal is to dynamically identify the devices using our verification strategy.

\begin{figure}[!ht]
    \centering
    \includegraphics[width=\linewidth]{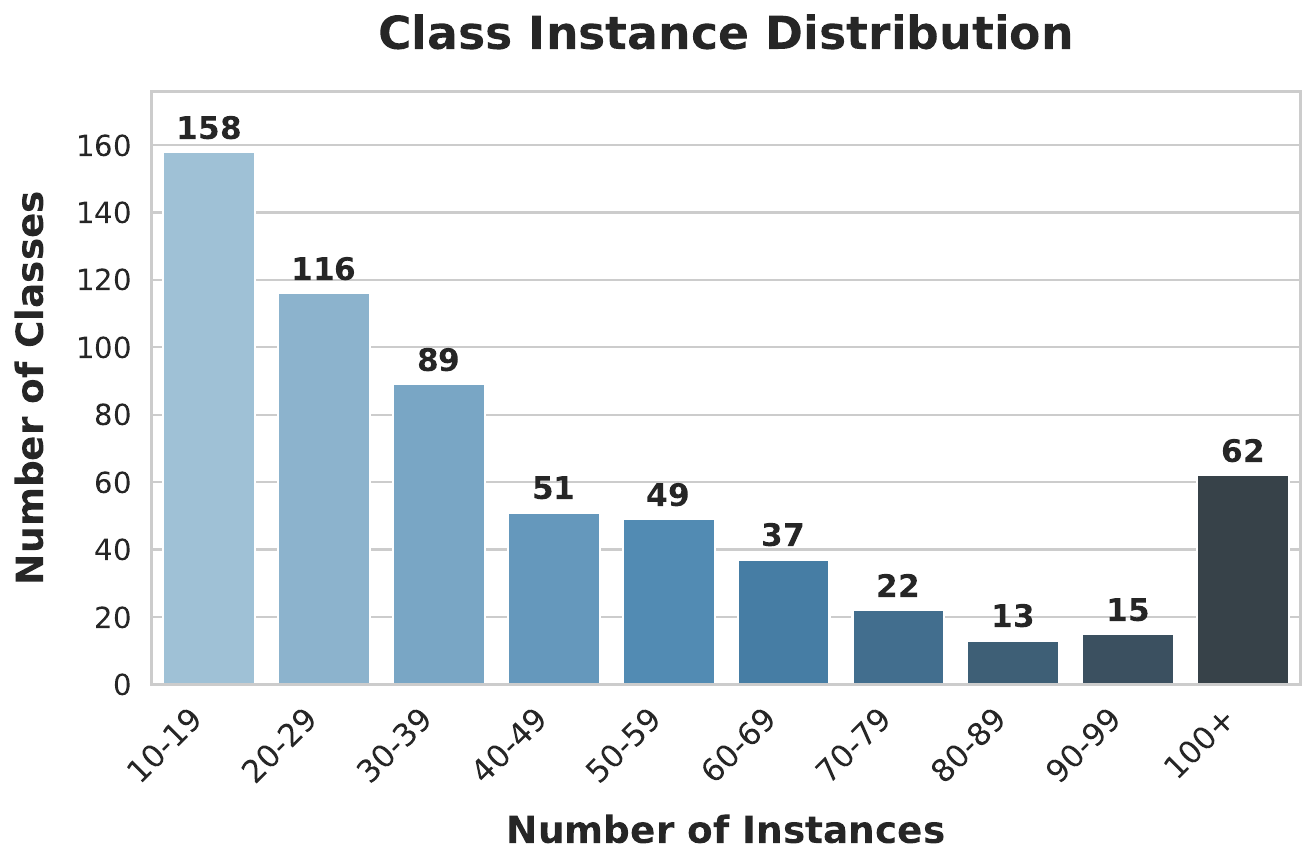}
    \caption{Distribution of the number of instances per class.}
    \label{fig:dev_dist}
\end{figure}

In order to ensure each class is sufficiently represented, we utilize the devices associated with ten or more images.
By applying this split, our working dataset is comprised of $33,668$ images represented across $612$ devices.
As shown in Figure~\ref{fig:dev_dist}, the dataset is very imbalanced with regard to the number of instances, with around 50\% of the dataset ($16,644$ images) being represented by 15\% of the classes ($90$ devices).

\section{Experiments}
\label{sec:experiments}

\begin{table}[!ht]
    \centering
    \caption{Number of instances per partition.}
    \begin{tabular}{lccc} \toprule
        \textbf{Experiment} & \textbf{Partition} & \textbf{Instances} & \textbf{Devices}  \\ \midrule
        \multirow{3}{*}{Intra-Device} & Training & $20200$ & $612$ \\
        & Validation &  $6734$ & $612$ \\
        & Testing &  $6734$ & $612$ \\ \midrule
        \multirow{3}{*}{Cross-Device} & Training &  $20914$ & $368$ \\
        & Validation &  $6477$  & $122$ \\
        & Testing &  $6277$ & $122$  \\ \bottomrule
    \end{tabular}
    \label{tab:stats}
\end{table}

In order to experimentally validate our approach, we devise intra-device and cross-device experimental scenarios.
For the intra-device scenario, we divide the images of each device chosen from the LPLCv2 dataset into training, validation and testing partitions in a \(60\)/\(20\)/\(20\) fashion.
This ensures that validation and testing are performed exclusively on unseen images from known devices.

For the cross-device scenario, we split the dataset using the same partitioning schema, but the separation is done at device level instead of image level.
This means that the known $612$ devices are divided into training, validation and testing such that all images from the same camera belong to only one partition exclusively.
Similarly, this ensures that validation and testing are performed exclusively on unknown devices.
The resulting statistics can be found in Table~\ref{tab:stats}.

Our experiments are then carried out in three steps.
In the first step, we train the model for verification using the Triplet loss and the dataset partitions previously defined.
Then, we generate a static set of $50,000$ random pairs of images from the testing set to serve as a baseline, keeping a ratio of $0.5$ between genuine and impostor pairs.
We then evaluate the verification model on the testing partition using the binary accuracy and F1-score metrics.
Then, we use the resulting feature extractor on the unknown class predictors defined in Section~\ref{sec:method}, using two baselines that represent the second and third experiment~steps.

The second step consists of evaluating each algorithm when a priori knowledge is available.
This means that the training set instances are available and their labels are known to the algorithm.
For the naive approach, this means that the initial model is initialized using the training set.
For the Leiden approach, the training instances are incorporated into the graph and the initial membership is provided, such that new instances are assigned to class $0$ and known instances are assigned to their corresponding classes.
Then, the third step consists of evaluating both approaches with no a priori information.

In both cases, we utilize the V-measure score with \(\beta = 1.0\) to evaluate the resulting labels.
This measure is appropriate for our task as it is independent of the label IDs, and provides an accurate measurement of zero-shot labeling efficiency.

Finally, we also use a PARSeq-tiny model~\cite{bautista2022parseq} on the OCR task on LPLCv2.
We choose PARSeq due to its common usage in recent ALPR literature and high performance on scenarios similar to the ones presented in LPLCv2~\cite{wojcik2025lplc,lima2026toward}.
We utilize the same splits, training it for intra- and cross-device scenarios in order to investigate the potential device biases in a real-world OCR task.
Here, we use the standard parameters and train both pre-trained and from scratch versions.

\subsection{Training Parameters}

\begin{table}[!ht]
    \centering
    \caption{Verification Hyperparameters}
    \begin{tabular}{lc} \toprule
        \textbf{Parameter} & \textbf{Value} \\ \midrule
        Epochs &  $1000$ \\
        Batches per Epoch & $120$ \\
        Early Stopping Patience & $40$ \\
        Early Stopping Metric & Validation Accuracy \\ \midrule
        Optimizer & Adam~\cite{kingma2015adam} \\
        Initial Learning Rate & $1e-4$ \\
        LR Factor & $0.75$ \\
        LR Patience & $5$ epochs \\
        Minimum LR & $1e-6$ \\

        \bottomrule
    \end{tabular}
    \label{tab:parameters_vef}
\end{table}

All images are resized to fit into a $224\times224$ square, keeping the aspect ratio, positioned in the middle and padded with gray pixels.
We train a ViT-b16~\cite{dosovitskiy2021vit} architecture from scratch on the verification task on a NVIDIA RTX 6000 GPU.
The model is trained using a triplet loss~\cite{hoffer2015triplet}, with dynamically generated training and validation triplets.
The hyperparameters of the verification training step can be found in Table~\ref{tab:parameters_vef}.

The naive algorithm utilizes a threshold of $0.6$ to assign an image to a known cluster.
The Leiden algorithm uses the Constant Potts Model (CPM) quality function, which is suitable to this task as several tightly knit communities are to be expected at high verification accuracy levels.
The resolution parameter is set at $0.8$.

\section{Results}
\label{sec:results}

\begin{table}[!ht]
    \centering
    \caption{Results for the verification baseline on the testing set.}
    \begin{tabular}{lcccc} \toprule
    \multirow{2}{*}{\textbf{Scenario}} & \multirow{2}{*}{\begin{tabular}[c]{@{}c@{}}\textbf{Accuracy} \\ ($\uparrow$)\end{tabular}} & \multirow{2}{*}{\begin{tabular}[c]{@{}c@{}}\textbf{Precision} \\ ($\uparrow$)\end{tabular}} & \multirow{2}{*}{\begin{tabular}[c]{@{}c@{}}\textbf{Recall} \\ ($\uparrow$)\end{tabular}} & \multirow{2}{*}{\begin{tabular}[c]{@{}c@{}}\textbf{F1-score} \\ ($\uparrow$)\end{tabular}} \\ \\ \midrule
    Intra-device & $98.13$ & $98.87$ & $97.36$ & $98.11$ \\
    Cross-device & $93.79$ & $97.83$ & $89.52$ & $93.45$ \\ \bottomrule
    \end{tabular}
    \label{tab:ver_results}
\end{table}

Table~\ref{tab:ver_results} presents the results for the verification task.
We report both Accuracy and F1-score on the testing partition of $50,000$ randomly generated pairs.
As expected, cross-device performance is significantly lower, which highlights the device bias present in the dataset.
The data is not homogeneous: a model trained on one partition does not necessarily generalize towards another partition.

In particular, we notice a higher incidence of false negatives, as seen in the lower Recall rate in the cross-device scenario.
This essentially means that, in verification, the features face a bigger hurdle when accurately distinguishing false negatives, that is, many genuine pairs are rejected.

\begin{table}[!ht]
    \centering
    \caption{Results for the Naive Algorithm.}
    \begin{tabular}{lccc} \toprule

    \multirow{2}{*}{\textbf{Scenario}} & \multirow{2}{*}{\textbf{Initialization}} & \multirow{2}{*}{\begin{tabular}[c]{@{}c@{}}\textbf{V-Measure} \\ ($\uparrow$)\end{tabular}} & \multirow{2}{*}{\begin{tabular}[c]{@{}c@{}}\textbf{Correct} \\ \textbf{Attributions} ($\uparrow$)\end{tabular}} \\  \\ \midrule
     
    \multirow{2}{*}{Intra-device} & None & $89.68$ & $64.52$ \\
     & Training & $92.30$ & $82.09$ \\ \midrule
    \multirow{2}{*}{Cross-device} & None & $73.45$ & $43.10$ \\  
     & Training &  $72.51$ & $33.90$ \\ \bottomrule
    \end{tabular}
    \label{tab:naive_results}
\end{table}

Table~\ref{tab:naive_results} presents our zero-shot results for the Naive algorithm.
We report the V-Measure and the instantaneous accuracy, meaning that for each incoming instance a correct attribution is recorded if either: the image is assigned to a new class and other images of the same device are not yet present in the known set, or, the image is assigned to an existing class in the known set and most of the same images from its corresponding device are also present in that class.

As the results show, a priori knowledge significantly increases performance for the intra-device scenario, whereas it worsens cross-device efficacy.
Indeed, the cross-device scenario is unable to leverage known information as its distribution is outside the range of the training set.
However, these results show that possessing the known information is often desirable for robust and successful executions of the~algorithm.

\begin{table}[!ht]
    \centering
    \caption{Results for the Leiden Algorithm (V-Measure).}
    \begin{tabular}{lcccc} \toprule
    \textbf{Scenario} & \textbf{Known} & \begin{tabular}[c]{@{}c@{}}\textbf{Homogeneity} \\ ($\uparrow$)\end{tabular} & \begin{tabular}[c]{@{}c@{}}\textbf{Completeness} \\ ($\uparrow$)\end{tabular} & \begin{tabular}[c]{@{}c@{}}\textbf{V-Measure} \\ ($\uparrow$)\end{tabular} \\ \midrule
     
     \multirow{2}{*}{\begin{tabular}[c]{@{}c@{}}Intra \\ Device\end{tabular}}  & None & $99.22$ & $81.11$ & $89.25$ \\
     & Training & $99.26$ & $81.17$ & $89.31$ \\ \midrule
    \multirow{2}{*}{\begin{tabular}[c]{@{}c@{}}Cross \\ Device\end{tabular}} & None & $99.18$ & $67.23$ & $80.14$ \\  
     & Training & $95.34$ & $63.66$ & $76.35$ \\ \bottomrule
    \end{tabular}
    \label{tab:leiden_results}
\end{table}

Table~\ref{tab:leiden_results} presents the results for the Leiden algorithm clustering.
The same trend seen in the Naive algorithm results persists, where a priori knowledge is helpful for classifying known devices, but harmful for new, unknown devices.
Our approach reaches a solid $80.14$ V-Measure score, highlighting its generalization potential.
This result is directly tied to the verification efficiency.
The goal when using node clustering on fuzzy relationships is to filter out the noise from erroneous verification steps.
While the implemented pipeline manages generalization, the performance gap can still be improved by refining the verification task performance.

A qualitative analysis of the results supports the intuition behind the good homogeneity but low completeness scores found in this scenario.
While there is little class contamination inside the same class, some classes are split up into many chunks.
This highlights the main limitation of the proposed method, which is a consequence of the relatively lower recall (a higher false-negative~rate).

\begin{table}[!ht]
    \centering
    \caption{PARSeq-tiny OCR Performance across Classes.}
    \begin{tabular}{llcccc} \toprule
        \multirow{3}{*}{\textbf{Scenario}} & \multirow{3}{*}{\textbf{Class}} & \multicolumn{2}{c}{\textbf{From Scratch}} & \multicolumn{2}{c}{\textbf{Pretrained}} \\ \cmidrule(lr){3-4} \cmidrule(lr){5-6}
        & & \begin{tabular}[c]{@{}c@{}}\textbf{Plate} \\ \textbf{Acc. (\%)}\end{tabular} & \begin{tabular}[c]{@{}c@{}}\textbf{Char.} \\ \textbf{Acc. (\%)}\end{tabular} & \begin{tabular}[c]{@{}c@{}}\textbf{Plate} \\ \textbf{Acc. (\%)}\end{tabular} & \begin{tabular}[c]{@{}c@{}}\textbf{Char.} \\ \textbf{Acc. (\%)}\end{tabular} \\ \midrule
        \multirow{5}{*}{\begin{tabular}[c]{@{}l@{}}Intra \\ Device\end{tabular}} 
        & Perfect   & $98.73$ & $99.72$ & $99.01$ & $99.78$ \\
        & Good   & $94.43$ & $98.72$ & $95.17$ & $98.96$ \\
        & Poor        & $84.82$ & $97.21$ & $87.12$ & $97.63$ \\
        & Illegible   & $65.91$ & $91.02$ & $68.18$ & $91.34$ \\ \cmidrule{2-6}
        & Overall & $95.29$ & $99.00$ & $95.98$ & $99.15$ \\ \midrule
        \multirow{5}{*}{\begin{tabular}[c]{@{}l@{}}Cross \\ Device\end{tabular}} 
        & Perfect   & $98.72$ & $99.78$ & $99.08$ & $99.84$ \\
        & Good   & $94.29$ & $98.94$ & $96.16$ & $99.25$ \\
        & Poor   & $84.52$ & $97.06$ & $85.60$ & $97.38$ \\
        & Illegible   & $56.06$ & $85.61$ & $50.76$ & $82.90$ \\ \cmidrule{2-6}
        & Overall & $93.41$ & $98.66$ & $94.26$ & $98.80$ \\ \bottomrule
    \end{tabular}
    \label{tab:ocr_results}
\end{table}

Finally, we present the OCR results with PARSeq on the partitions defined for our experiments in this paper, with the results stratified according to the plate-wise legibility labels from the dataset.
These are presented in Table~\ref{tab:ocr_results}.
While the pre-trained model leverages its training data in order to achieve a robust generalization performance, the model trained from scratch reveals the device bias present in the dataset.
In the cross-device scenario, the whole-plate accuracy drops from $95.29$ to $93.41$, a drop of $1.88$ percentage points that represents a significant increase in error rate.
This is especially true for low-resolution, poorly readable license plates, with Illegible plates representing the most impacted class.

\section{Conclusion}
\label{sec:conclusion}

In this paper, we have presented a novel method for zero-shot classification based on the verification task.
Drawing influence from the human-based process of recognition of new data, as well as from the state of the art in facial recognition, we devise a preliminary method for automatic new class detection.
We employ both a naive algorithm, which assumes verification accuracy is perfect, and a modular Leiden-based graph clustering for class (community) detection, and show the results for the LPLCv2 dataset.

Although the initial results seem promising, several challenges remain.
Our results show that these approaches often fail at recognizing the lesser represented classes, even when constrained to a subset with at least ten instances per class.
Also, the verification network often fails at separating similar classes, which introduces a fair amount of noise in the zero-shot step.
Finally, the zero-shot method does not yet reach a satisfactory performance, which highlights the need for future~research.

Future work will focus on improving the naive approach to exploit new incoming information at test time and correct past mistakes, leveraging the high internal cohesion of each class that is presumed at high verification performance levels.
Also, an important step is to refine the quality of the extracted features by improving inter-class separation for similar classes.
Finally, another future direction for research is to expand the work's scope towards novel features of other datasets in order to validate the approach in more general scenarios.

\section*{Acknowledgment}

This study was financed in part by the \textit{Coordenação de Aperfeiçoamento de Pessoal de Nível Superior - Brasil~(CAPES)}, through the \textit{Programa de Excelência Acadêmica~(PROEX)} - Finance Code 001, in part by the \textit{Fundação Araucária} under grant \#~078/2026, and in part by the \textit{Conselho Nacional de Desenvolvimento Científico e Tecnológico~(CNPq)} (\#~315409/2023-1).

\bibliographystyle{sbc}
\bibliography{example}

@misc{wojcik2026lplcv2,
      title={LPLCv2: An Expanded Dataset for Fine-Grained License Plate Legibility Classification}, 
      author = {L. {Wojcik} and E. A. F. {Machoski} and E. {Nascimento Jr.} and R. {Laroca} and D. {Menotti}},
      year={2026},
      eprint={2604.08741},
      archivePrefix={arXiv},
      primaryClass={cs.CV},
      url={https://arxiv.org/abs/2604.08741}, 
}

@article{wojcik2025lplc,
  title = {{LPLC}: A Dataset for License Plate Legibility Classification},
  author = {L. {Wojcik} and G. E. {Lima} and V. {Nascimento} and E. {Nascimento Jr.} and R. {Laroca} and D. {Menotti}},
  year = {2025},
  journal = {Conference on Graphics, Patterns and Images (SIBGRAPI)},
  volume = {},
  number = {},
  pages = {1-6},
  doi = {10.1109/SIBGRAPI67909.2025.11223367},
  issn = {1530-1834},
}

@Article{vonluxburg2007spectral,
author={von Luxburg, Ulrike},
title={A tutorial on spectral clustering},
journal={Statistics and Computing},
year={2007},
month={Dec},
day={01},
volume={17},
number={4},
pages={395-416},
issn={1573-1375},
doi={10.1007/s11222-007-9033-z},
url={https://doi.org/10.1007/s11222-007-9033-z}
}

@ARTICLE{khalid2023facial,
  author={Khalid, Syed Safwan and Awais, Muhammad and Feng, Zhen-Hua and Chan, Chi-Ho and Farooq, Ammarah and Akbari, Ali and Kittler, Josef},
  journal={IEEE Transactions on Pattern Analysis and Machine Intelligence}, 
  title={NPT-Loss: Demystifying Face Recognition Losses With Nearest Proxies Triplet}, 
  year={2023},
  volume={45},
  number={12},
  pages={15249-15259},
  doi={10.1109/TPAMI.2022.3162705}}

@article{du2022elements,
  title={The elements of end-to-end deep face recognition: A survey of recent advances},
  author={Du, Hang and Shi, Hailin and Zeng, Dan and Zhang, Xiao-Ping and Mei, Tao},
  journal={ACM computing surveys (CSUR)},
  volume={54},
  number={10s},
  pages={1--42},
  year={2022},
  publisher={ACM New York, NY}
}

@article{liu2020energy,
      title={Energy-based Out-of-distribution Detection},
      author={Liu, Weitang and Wang, Xiaoyun and Owens, John and Li, Yixuan},
      journal={Advances in Neural Information Processing Systems},
      year={2020}
}

@article{lu2025out,
  title={Out-of-distribution detection: A task-oriented survey of recent advances},
  author={Lu, Shuo and Wang, Yingsheng and Sheng, Lijun and He, Lingxiao and Zheng, Aihua and Liang, Jian},
  journal={ACM Computing Surveys},
  volume={58},
  number={2},
  pages={1--39},
  year={2025},
  publisher={ACM New York, NY}
}

@InProceedings{mensink2014costa,
author = {Mensink, Thomas and Gavves, Efstratios and Snoek, Cees G.M.},
title = {COSTA: Co-Occurrence Statistics for Zero-Shot Classification},
booktitle = {Proceedings of the IEEE Conference on Computer Vision and Pattern Recognition (CVPR)},
month = {June},
year = {2014}
}

@Article{traag2019leiden,
author={Traag, V. A.
and Waltman, L.
and van Eck, N. J.},
title={From Louvain to Leiden: guaranteeing well-connected communities},
journal={Scientific Reports},
year={2019},
month={Mar},
day={26},
volume={9},
number={1},
pages={5233},
issn={2045-2322},
doi={10.1038/s41598-019-41695-z},
url={https://doi.org/10.1038/s41598-019-41695-z}
}

@InProceedings{vaze2022gcd,
               title={Generalized Category Discovery},
               author={Sagar Vaze and Kai Han and Andrea Vedaldi and Andrew Zisserman},
               booktitle={IEEE Conference on Computer Vision and Pattern Recognition},
               year={2022}}

@inproceedings{wen2023parametric,
  title={Parametric classification for generalized category discovery: A baseline study},
  author={Wen, Xin and Zhao, Bingchen and Qi, Xiaojuan},
  booktitle={2023 IEEE/CVF International Conference on Computer Vision (ICCV)},
  pages={16544--16554},
  year={2023},
  organization={IEEE}
}

@article{zha2025data,
  title={Data-centric artificial intelligence: A survey},
  author={Zha, Daochen and Bhat, Zaid Pervaiz and Lai, Kwei-Herng and Yang, Fan and Jiang, Zhimeng and Zhong, Shaochen and Hu, Xia},
  journal={ACM Computing Surveys},
  volume={57},
  number={5},
  pages={1--42},
  year={2025},
  publisher={ACM New York, NY}
}

@inproceedings{hoffmann2022training,
author = {Hoffmann, Jordan and others},
title = {Training compute-optimal large language models},
year = {2022},
isbn = {9781713871088},
publisher = {Curran Associates Inc.},
address = {Red Hook, NY, USA},
booktitle = {Proceedings of the 36th International Conference on Neural Information Processing Systems},
articleno = {2176},
numpages = {15},
location = {New Orleans, LA, USA},
series = {NIPS '22}
}

@inproceedings{laroca2022first,
  title = {A First Look at Dataset Bias in License Plate Recognition},
  author = {R. {Laroca} and M. {Santos} and V. {Estevam} and E. {Luz} and D. {Menotti}},
  year = {2022},
  month = {Oct},
  booktitle = {Conference on Graphics, Patterns and Images (SIBGRAPI)},
  volume = {},
  number = {},
  pages = {234-239},
  doi = {10.1109/SIBGRAPI55357.2022.9991768},
  issn = {1530-1834}
}

@inproceedings{hendrycks2017baseline,
  title={A Baseline for Detecting Misclassified and Out-of-Distribution Examples in Neural Networks},
  author={Dan Hendrycks and Kevin Gimpel},
  booktitle={International Conference on Learning Representations},
  year={2017},
  url={https://openreview.net}
}

@inproceedings{kingma2015adam,
  author    = {Kingma, Diederik P. and Ba, Jimmy},
  title     = {Adam: A Method for Stochastic Optimization},
  booktitle = {International Conference on Learning Representations (ICLR)},
  year      = {2015},
  url       = {https://arxiv.org/abs/1412.6980}
}

@InProceedings{hoffer2015triplet,
author="Hoffer, Elad
and Ailon, Nir",
editor="Feragen, Aasa
and Pelillo, Marcello
and Loog, Marco",
title="Deep Metric Learning Using Triplet Network",
booktitle="Similarity-Based Pattern Recognition",
year="2015",
publisher="Springer International Publishing",
address="Cham",
pages="84--92",
isbn="978-3-319-24261-3"
}

@article{felipe2025leiden,
  title={From Leiden to Pleasure Island: The Constant Potts Model for community detection as a hedonic game},
  author={Felipe, Lucas Lopes and Avrachenkov, Konstantin and Menasch{\'e}, Daniel Sadoc},
  journal={Physica A: Statistical Mechanics and its Applications},
  pages={130989},
  year={2025},
  publisher={Elsevier}
}

@InProceedings{bautista2022parseq,
  title={Scene Text Recognition with Permuted Autoregressive Sequence Models},
  author={Bautista, Darwin and Atienza, Rowel},
  booktitle={European Conference on Computer Vision},
  pages={178--196},
  month={10},
  year={2022},
  publisher={Springer Nature Switzerland},
  address={Cham},
  doi={10.1007/978-3-031-19815-1_11},
  url={https://doi.org/10.1007/978-3-031-19815-1_11}
}

@inproceedings{dosovitskiy2021vit,
  title = {An Image is Worth 16x16 Words: Transformers for Image Recognition at Scale},
  author = {Alexey {Dosovitskiy} and others},
  year = 2021,
  booktitle = {International Conference on Learning Representations~(ICLR)},
  volume = {},
  number = {},
  pages = {1--22}
}

@InProceedings{xian2016latent,
author = {Xian, Yongqin and Akata, Zeynep and Sharma, Gaurav and Nguyen, Quynh and Hein, Matthias and Schiele, Bernt},
title = {Latent Embeddings for Zero-Shot Classification},
booktitle = {Proceedings of the IEEE Conference on Computer Vision and Pattern Recognition (CVPR)},
month = {June},
year = {2016}
}

@InProceedings{novack2023chils,
  title = 	 {{CH}i{LS}: Zero-Shot Image Classification with Hierarchical Label Sets},
  author =       {Novack, Zachary and Mcauley, Julian and Lipton, Zachary Chase and Garg, Saurabh},
  booktitle = 	 {Proceedings of the 40th International Conference on Machine Learning},
  pages = 	 {26342--26362},
  year = 	 {2023},
  editor = 	 {Krause, Andreas and Brunskill, Emma and Cho, Kyunghyun and Engelhardt, Barbara and Sabato, Sivan and Scarlett, Jonathan},
  volume = 	 {202},
  series = 	 {Proceedings of Machine Learning Research},
  month = 	 {23--29 Jul},
  publisher =    {PMLR},
  url = 	 {https://proceedings.mlr.press/v202/novack23a.html}
}

@INPROCEEDINGS{saha2024text,
  author={Saha, Oindrila and Van Horn, Grant and Maji, Subhransu},
  booktitle={2024 IEEE/CVF Conference on Computer Vision and Pattern Recognition (CVPR)}, 
  title={Improved Zero-Shot Classification by Adapting VLMs with Text Descriptions}, 
  year={2024},
  volume={},
  number={},
  pages={17542-17552},
  doi={10.1109/CVPR52733.2024.01661}}

@INPROCEEDINGS{shipard2023synthetic,
  author={Shipard, Jordan and Wiliem, Arnold and Thanh, Kien Nguyen and Xiang, Wei and Fookes, Clinton},
  booktitle={2023 IEEE/CVF Conference on Computer Vision and Pattern Recognition Workshops (CVPRW)}, 
  title={Diversity is Definitely Needed: Improving Model-Agnostic Zero-shot Classification via Stable Diffusion}, 
  year={2023},
  volume={},
  number={},
  pages={769-778},
  doi={10.1109/CVPRW59228.2023.00084}}

@article{ling2025domain,
  title={Domain specialization as the key to make large language models disruptive: A comprehensive survey},
  author={Ling, Chen and Zhao, Xujiang and Lu, Jiaying and Deng, Chengyuan and Zheng, Can and Wang, Junxiang and Chowdhury, Tanmoy and Li, Yun and Cui, Hejie and Zhang, Xuchao and others},
  journal={ACM Computing Surveys},
  volume={58},
  number={3},
  pages={1--39},
  year={2025},
  publisher={ACM New York, NY}
}

@inproceedings{zhang2024generation,
  title={Generation-driven contrastive self-training for zero-shot text classification with instruction-following LLM},
  author={Zhang, Ruohong and Wang, Yau-Shian and Yang, Yiming},
  booktitle={Proceedings of the 18th Conference of the European Chapter of the Association for Computational Linguistics (Volume 1: Long Papers)},
  pages={659--673},
  year={2024}
}

@inproceedings{yin2019benchmarking,
    title = "Benchmarking Zero-shot Text Classification: Datasets, Evaluation and Entailment Approach",
    author = "Yin, Wenpeng  and
      Hay, Jamaal  and
      Roth, Dan",
    editor = "Inui, Kentaro  and
      Jiang, Jing  and
      Ng, Vincent  and
      Wan, Xiaojun",
    booktitle = "Proceedings of the 2019 Conference on Empirical Methods in Natural Language Processing and the 9th International Joint Conference on Natural Language Processing (EMNLP-IJCNLP)",
    month = nov,
    year = "2019",
    address = "Hong Kong, China",
    publisher = "Association for Computational Linguistics",
    url = "https://aclanthology.org/D19-1404/",
    doi = "10.18653/v1/D19-1404",
    pages = "3914--3923"
}

@INPROCEEDINGS{zhang2023decoupling,
  author={Zhang, Zihan and Xiang, Xiang},
  booktitle={2023 IEEE/CVF Conference on Computer Vision and Pattern Recognition (CVPR)}, 
  title={Decoupling MaxLogit for Out-of-Distribution Detection}, 
  year={2023},
  volume={},
  number={},
  pages={3388-3397},
  doi={10.1109/CVPR52729.2023.00330}}

@Article{yang2024survey,
author={Yang, Jingkang
and Zhou, Kaiyang
and Li, Yixuan
and Liu, Ziwei},
title={Generalized Out-of-Distribution Detection: A Survey},
journal={International Journal of Computer Vision},
year={2024},
month={Dec},
day={01},
volume={132},
number={12},
pages={5635-5662},
issn={1573-1405},
doi={10.1007/s11263-024-02117-4},
url={https://doi.org/10.1007/s11263-024-02117-4}
}

@article{lima2026toward,
  title = {Toward Unified Fine-Grained Vehicle Classification and Automatic License Plate Recognition},
  author = {G. E. {Lima} and V. {Nascimento} and E. {Santos} and E. {Nascimento Jr.} and R. {Laroca} and D. {Menotti}},
  year = {2026},
  journal = {Journal of the Brazilian Computer Society},
  volume = {32},
  number = {1},
  pages = {783-799},
  doi = {10.5753/jbcs.2026.5899},
  issn = {},
}

\end{document}